# Inverse Covariance Estimation for High-Dimensional Data in Linear Time and Space: Spectral Methods for Riccati and Sparse Models

**Jean Honorio**
CSAIL, MIT
Cambridge, MA 02139, USA
jhonorio@csail.mit.edu

**Tommi Jaakkola**
CSAIL, MIT
Cambridge, MA 02139, USA
tommi@csail.mit.edu

## Abstract

We propose maximum likelihood estimation for learning Gaussian graphical models with a Gaussian ($\ell_2^2$) prior on the parameters. This is in contrast to the commonly used Laplace ($\ell_1$) prior for encouraging sparseness. We show that our optimization problem leads to a Riccati matrix equation, which has a closed form solution. We propose an efficient algorithm that performs a singular value decomposition of the training data. Our algorithm is $\mathcal{O}(NT^2)$-time and $\mathcal{O}(NT)$-space for $N$ variables and $T$ samples. Our method is tailored to high-dimensional problems ($N \gg T$), in which sparseness promoting methods become intractable. Furthermore, instead of obtaining a single solution for a specific regularization parameter, our algorithm finds the whole solution path. We show that the method has logarithmic sample complexity under the spiked covariance model. We also propose sparsification of the dense solution with provable performance guarantees. We provide techniques for using our learnt models, such as removing unimportant variables, computing likelihoods and conditional distributions. Finally, we show promising results in several gene expressions datasets.

## 1 Introduction

Estimation of large inverse covariance matrices, particularly when the number of variables $N$ is significantly larger than the number of samples $T$, has attracted increased attention recently. One of the main reasons for this interest is the need for researchers to discover interactions between variables in high dimensional datasets, in areas such as genetics, neuroscience and meteorology. For instance in gene expression datasets, $N$ is in the order of 20 thousands to 2 millions, while $T$ is in the order of few tens to few hundreds. Inverse covariance (precision) matrices are the natural parameterization of Gaussian graphical models.

In this paper, we propose maximum likelihood estimation for learning Gaussian graphical models with a Gaussian ($\ell_2^2$) prior on the parameters. This is in contrast to the commonly used Laplace ($\ell_1$) prior for encouraging sparseness. We consider the computational aspect of this problem, under the assumption that the number of variables $N$ is significantly larger than the number of samples $T$, i.e. $N \gg T$.

Our technical contributions in this paper are the following. First, we show that our optimization problem leads to a Riccati matrix equation, which has a closed form solution. Second, we propose an efficient algorithm that performs a singular value decomposition of the sample covariance matrix, which can be performed very efficiently through singular value decomposition of the training data. Third, we show logarithmic sample complexity of the method under the spiked covariance model. Fourth, we propose sparsification of the dense solution with provable performance guarantees. That is, there is a bounded degradation of the Kullback-Leibler divergence and expected log-likelihood. Finally, we provide techniques for using our learnt models, such as removing unimportant variables, computing likelihoods and conditional distributions.

## 2 Background

In this paper, we use the notation in Table 1.

A *Gaussian graphical model* is a graph in which all random variables are continuous and jointly Gaussian. This model corresponds to the multivariate normal distribution for $N$ variables with mean $\boldsymbol{\mu}$ and covariance matrix $\boldsymbol{\Sigma} \in \mathbb{R}^{N\times N}$. Conditional independence in a

Table 1: Notation used in this paper.

| Notation | Description |
|---|---|
| $\mathbf{A} \succeq \mathbf{0}$ | $\mathbf{A} \in \mathbb{R}^{N\times N}$ is symmetric and positive semidefinite |
| $\mathbf{A} \succ \mathbf{0}$ | $\mathbf{A} \in \mathbb{R}^{N\times N}$ is symmetric and positive definite |
| $\|\mathbf{A}\|_1$ | $\ell_1$-norm of $\mathbf{A} \in \mathbb{R}^{N\times M}$, i.e. $\sum_{nm} \|a_{nm}\|$ |
| $\|\mathbf{A}\|_\infty$ | $\ell_\infty$-norm of $\mathbf{A} \in \mathbb{R}^{N\times M}$, i.e. $\max_{nm} \|a_{nm}\|$ |
| $\|\mathbf{A}\|_2$ | spectral norm of $\mathbf{A} \in \mathbb{R}^{N\times N}$, i.e. the maximum eigenvalue of $\mathbf{A} \succ \mathbf{0}$ |
| $\|\mathbf{A}\|_{\mathfrak{F}}$ | Frobenius norm of $\mathbf{A} \in \mathbb{R}^{N\times M}$, i.e. $\sqrt{\sum_{nm} a_{nm}^2}$ |
| $\mathrm{tr}(\mathbf{A})$ | trace of $\mathbf{A} \in \mathbb{R}^{N\times N}$, i.e. $\mathrm{tr}(\mathbf{A}) = \sum_n a_{nn}$ |
| $\langle \mathbf{A}, \mathbf{B} \rangle$ | scalar product of $\mathbf{A}, \mathbf{B} \in \mathbb{R}^{N\times M}$, i.e. $\sum_{nm} a_{nm} b_{nm}$ |
| $\mathbf{A}\#\mathbf{B}$ | geometric mean of $\mathbf{A}, \mathbf{B} \in \mathbb{R}^{N\times N}$, i.e. the matrix $\mathbf{Z} \succeq \mathbf{0}$ with maximum singular values such that $\begin{bmatrix} \mathbf{A} & \mathbf{Z} \\ \mathbf{Z} & \mathbf{B} \end{bmatrix} \succeq \mathbf{0}$ |

Gaussian graphical model is simply reflected in the zero entries of the precision matrix $\mathbf{\Omega} = \mathbf{\Sigma}^{-1}$ (Lauritzen, 1996). Let $\mathbf{\Omega} = \{\omega_{n_1 n_2}\}$, two variables $n_1$ and $n_2$ are conditionally independent if and only if $\omega_{n_1 n_2} = 0$.

The log-likelihood of a sample $\mathbf{x} \in \mathbb{R}^N$ in a Gaussian graphical model with mean $\boldsymbol{\mu}$ and precision matrix $\mathbf{\Omega}$ is given by:

$$\mathcal{L}(\mathbf{\Omega}, \mathbf{x}) \equiv \log\det\mathbf{\Omega} - (\mathbf{x} - \boldsymbol{\mu})^\top \mathbf{\Omega} (\mathbf{x} - \boldsymbol{\mu}) \quad (1)$$

In this paper, we use a short hand notation for the average log-likelihood of $T$ samples $\mathbf{x}^{(1)}, \ldots, \mathbf{x}^{(T)} \in \mathbb{R}^N$. Given that this average log-likelihood depends only on the sample covariance matrix $\widehat{\mathbf{\Sigma}}$, we define:

$$\begin{aligned} \mathcal{L}(\mathbf{\Omega}, \widehat{\mathbf{\Sigma}}) &\equiv \frac{1}{T} \sum_t \mathcal{L}(\mathbf{\Omega}, \mathbf{x}^{(t)}) \\ &= \log\det\mathbf{\Omega} - \langle \widehat{\mathbf{\Sigma}}, \mathbf{\Omega} \rangle \end{aligned} \quad (2)$$

A very well known technique for the estimation of precision matrices is Tikhonov regularization (Duchi et al., 2008). Given a sample covariance matrix $\widehat{\mathbf{\Sigma}} \succeq \mathbf{0}$, the *Tikhonov-regularized problem* is defined as:

$$\max_{\mathbf{\Omega} \succ \mathbf{0}} \mathcal{L}(\mathbf{\Omega}, \widehat{\mathbf{\Sigma}}) - \rho \, \mathrm{tr}(\mathbf{\Omega}) \quad (3)$$

for regularization parameter $\rho > 0$. The term $\mathcal{L}(\mathbf{\Omega}, \widehat{\mathbf{\Sigma}})$ is the Gaussian log-likelihood as defined in eq.(2). The optimal solution of eq.(3) is given by:

$$\widehat{\mathbf{\Omega}} = (\widehat{\mathbf{\Sigma}} + \rho \mathbf{I})^{-1} \quad (4)$$

The estimation of sparse precision matrices was first introduced in (Dempster, 1972). It is well known that finding the most sparse precision matrix which fits a dataset is a NP-hard problem (Banerjee et al., 2006; Banerjee et al., 2008). Since the $\ell_1$-norm is the tighest convex upper bound of the cardinality of a matrix, several $\ell_1$-regularization methods have been proposed. Given a dense sample covariance matrix $\widehat{\mathbf{\Sigma}} \succeq \mathbf{0}$, the $\ell_1$*-regularized problem* is defined as:

$$\max_{\mathbf{\Omega} \succ \mathbf{0}} \mathcal{L}(\mathbf{\Omega}, \widehat{\mathbf{\Sigma}}) - \rho \|\mathbf{\Omega}\|_1 \quad (5)$$

for regularization parameter $\rho > 0$. The term $\mathcal{L}(\mathbf{\Omega}, \widehat{\mathbf{\Sigma}})$ is the Gaussian log-likelihood as defined in eq.(2). The term $\|\mathbf{\Omega}\|_1$ encourages sparseness of the precision matrix.

From a Bayesian viewpoint, eq.(5) is equivalent to maximum likelihood estimation with a prior that assumes that each entry in $\mathbf{\Omega}$ is independent and each of them follow a Laplace distribution ($\ell_1$-norm).

Several algorithms have been proposed for solving eq.(5): sparse regression for each variable (Meinshausen & Bühlmann, 2006; Zhou et al., 2011), determinant maximization with linear inequality constraints (Yuan & Lin, 2007), a block coordinate descent method with quadratic programming (Banerjee et al., 2006; Banerjee et al., 2008) or sparse regression (Friedman et al., 2007), a Cholesky decomposition approach (Rothman et al., 2008), a projected gradient method (Duchi et al., 2008), a projected quasi-Newton method (Schmidt et al., 2009), Nesterov's smooth optimization technique (Lu, 2009), an alternating linearization method (Scheinberg et al., 2010), a greedy coordinate descent method (Scheinberg & Rish, 2010), a block coordinate gradient descent method (Yun et al., 2011), quadratic approximation (Hsieh et al., 2011), Newton-like methods (Olsen et al., 2012), iterative thresholding (Guillot et al., 2012), greedy forward and backward steps (Johnson et al., 2012) and divide and conquer (Hsieh et al., 2012).

Additionally, a general rule that uses the sample covariance $\widehat{\mathbf{\Sigma}}$ and splits the graph into connected components was proposed in (Mazumder & Hastie, 2012). After applying this rule, one can use any of the above methods for each component separately. This technique enables the use of $\ell_1$-regularization methods to very high dimensional datasets. To the best of our knowledge, (Mazumder & Hastie, 2012) reported results on the highest dimensional dataset (24,481 genes).

## 3 Problem Setup

In this paper, we assume that the number of variables $N$ is significantly larger than the number of samples $T$, i.e. $N \gg T$. This is a reasonable assumption is sev-eral contexts, for instance in gene expression datasets where $N$ is in the order of 20 thousands to 2 millions, while $T$ is in the order of few tens to few hundreds.

Given a sample covariance matrix $\widehat{\mathbf{\Sigma}} \succeq \mathbf{0}$, we define the *Riccati-regularized problem* as:

$$\max_{\mathbf{\Omega} \succ \mathbf{0}} \mathcal{L}(\mathbf{\Omega}, \widehat{\mathbf{\Sigma}}) - \frac{\rho}{2} \|\mathbf{\Omega}\|_{\mathfrak{F}}^2 \tag{6}$$

for regularization parameter $\rho > 0$. The term $\mathcal{L}(\mathbf{\Omega}, \widehat{\mathbf{\Sigma}})$ is the Gaussian log-likelihood as defined in eq.(2). The term $\|\mathbf{\Omega}\|_{\mathfrak{F}}$ is the Frobenius norm. We chose the name "Riccati regularizer" because the optimal solution of eq.(6) leads to a *Riccati matrix equation.*

From a Bayesian viewpoint, eq.(6) is equivalent to maximum likelihood estimation with a prior that assumes that each entry in $\mathbf{\Omega}$ is independent and each of them follow a Gaussian distribution ($\ell_2^2$-norm).

Surprisingly, the problem in eq.(6) has not received much attention. The problem was briefly mentioned in (Witten & Tibshirani, 2009) as a subproblem of a *regression* technique. A tangentially related prior is the $\ell_2^2$-norm regularization of the Cholesky factors of the precision matrix (Huang et al., 2006).

It is well known that $\ell_1$-regularization allows obtaining an inverse covariance matrix that is "simple" in the sense that it is sparse, with a small number of non-zero entries. Our proposed $\ell_2^2$-regularizer allows obtaining an inverse covariance that is also "simple" in the sense that it is low-rank as we will show in Section 4. On the other hand, the solution of the $\ell_1$-regularized problem in eq.(5) is full-rank with $N$ different eigenvalues, even for simple cases such as datasets consisting of just two samples. One seemingly unavoidable bottleneck for $\ell_1$-regularization is the computation of the covariance matrix $\widehat{\mathbf{\Sigma}} \in \mathbb{R}^{N \times N}$ which is $\mathcal{O}(N^2 T)$-time and $O(N^2)$-space. Moreover, the work of (Banerjee et al., 2006) showed that in order to obtain an $\varepsilon$-accurate solution, we need $\mathcal{O}(N^{4.5}/\varepsilon)$-time. It is also easy to argue that without further assumptions, sparseness promoting algorithms are $\Omega(N^3)$-time and $O(N^2)$-space. This is prohibitive for very high dimensional datasets. The method of connected components (Mazumder & Hastie, 2012) can reduce the requirement of $\Omega(N^3)$-time. Unfortunately, in order to make the biggest connected component of a reasonable size $N' = 500$ (as prescribed in (Mazumder & Hastie, 2012)), a very large regularization parameter $\rho$ is needed. As a result, the learnt models do not generalize well as we will show experimentally in Section 5.

# 4 Linear-Time and Space Algorithms

In this section, we propose algorithms with time-complexity $\mathcal{O}(NT^2)$ and space-complexity $\mathcal{O}(NT)$ by using a low-rank parameterization. Given our assumption that $N \gg T$, we can say that our algorithms are linear time and space, i.e. $\mathcal{O}(N)$.

## 4.1 Solution of the Riccati Problem

Here we show that the solution of our proposed Riccati-regularized problem leads to the Riccati matrix equation, which has a closed form solution.

**Theorem 1.** *For $\rho > 0$, the optimal solution of the* Riccati-regularized problem *in eq.(6) is given by:*

$$\widehat{\mathbf{\Omega}} = \lim_{\varepsilon \to 0^+} \left( \left( \frac{1}{\rho} \widehat{\mathbf{\Sigma}}_\varepsilon \right) \# \left( \widehat{\mathbf{\Sigma}}_\varepsilon^{-1} + \frac{1}{4\rho} \widehat{\mathbf{\Sigma}}_\varepsilon \right) - \frac{1}{2\rho} \widehat{\mathbf{\Sigma}}_\varepsilon \right) \tag{7}$$

*where $\widehat{\mathbf{\Sigma}}_\varepsilon = \widehat{\mathbf{\Sigma}} + \varepsilon \mathbf{I}$.*

*Proof.* First, we consider $\widehat{\mathbf{\Sigma}}_\varepsilon = \widehat{\mathbf{\Sigma}} + \varepsilon \mathbf{I} \succ \mathbf{0}$ for some small $\varepsilon > 0$ instead of $\widehat{\mathbf{\Sigma}} \succeq \mathbf{0}$. The $\varepsilon$-corrected version of eq.(6), that is $\max_{\mathbf{\Omega} \succ \mathbf{0}} \mathcal{L}(\mathbf{\Omega}, \widehat{\mathbf{\Sigma}}_\varepsilon) - \frac{\rho}{2} \|\mathbf{\Omega}\|_{\mathfrak{F}}^2$ has the minimizer $\widehat{\mathbf{\Omega}}_\varepsilon$ if and only if the derivative of the objective function at $\mathbf{\Omega} = \widehat{\mathbf{\Omega}}_\varepsilon$ is equal to zero. That is, $\widehat{\mathbf{\Omega}}_\varepsilon^{-1} - \widehat{\mathbf{\Sigma}}_\varepsilon - \rho \widehat{\mathbf{\Omega}}_\varepsilon = \mathbf{0}$. The latter equation is known as the *Riccati matrix equation*, which by virtue of Theorem 3.1 in (Lim, 2006) has the solution $\widehat{\mathbf{\Omega}}_\varepsilon = \left( \frac{1}{\rho} \widehat{\mathbf{\Sigma}}_\varepsilon \right) \# \left( \widehat{\mathbf{\Sigma}}_\varepsilon^{-1} + \frac{1}{4\rho} \widehat{\mathbf{\Sigma}}_\varepsilon \right) - \frac{1}{2\rho} \widehat{\mathbf{\Sigma}}_\varepsilon$. By taking the limit $\varepsilon \to 0^+$, we prove our claim. □

Although the optimal solution in eq.(7) seems to require $\Omega(N^2)$ and $\mathcal{O}(N^3)$ operations, in the next section we show efficient algorithms when $N \gg T$.

## 4.2 Spectral Method for the Riccati and Tikhonov Problems

In what follows, we provide a method for computing the optimal solution of the Riccati-regularized problem as well as the Tikhonov-regularized problem by using singular value decomposition of the training data. First, we focus on the Riccati-regularized problem.

**Theorem 2.** *Let $\widehat{\mathbf{\Sigma}} = \mathbf{U}\mathbf{D}\mathbf{U}^\top$ be the singular value decomposition of $\widehat{\mathbf{\Sigma}}$, where $\mathbf{U} \in \mathbb{R}^{N \times T}$ is an orthonormal matrix (i.e. $\mathbf{U}^\top \mathbf{U} = \mathbf{I}$) and $\mathbf{D} \in \mathbb{R}^{T \times T}$ is a diagonal matrix. For $\rho > 0$, the optimal solution of the* Riccati-regularized problem *in eq.(6) is given by:*

$$\widehat{\mathbf{\Omega}} = \mathbf{U}\widetilde{\mathbf{D}}\mathbf{U}^\top + \frac{1}{\sqrt{\rho}} \mathbf{I} \tag{8}$$

*where $\widetilde{\mathbf{D}} \in \mathbb{R}^{T \times T}$ is a diagonal matrix with entries $(\forall t)$ $\widetilde{d}_{tt} = \sqrt{\frac{1}{\rho} + \frac{d_{tt}^2}{4\rho^2}} - \frac{d_{tt}}{2\rho} - \frac{1}{\sqrt{\rho}}$.*

*Proof Sketch.* We consider an *over-complete* singular value decomposition by adding columns to $\mathbf{U}$, thus obtaining $\bar{\mathbf{U}} \in \mathbb{R}^{N \times N}$ such that $\bar{\mathbf{U}}\bar{\mathbf{U}}^\top = \mathbf{I}$, which allows obtaining a singular value decomposition for $\widehat{\mathbf{\Sigma}}_\varepsilon$. Then we apply Theorem 1 and properties of the geometric mean for general as well as diagonal matrices. $\square$

(Please, see Appendix B for detailed proofs.)

Next, we focus on the Tikhonov-regularized problem.

**Theorem 3.** *Let $\widehat{\mathbf{\Sigma}} = \mathbf{U}\mathbf{D}\mathbf{U}^\top$ be the singular value decomposition of $\widehat{\mathbf{\Sigma}}$, where $\mathbf{U} \in \mathbb{R}^{N\times T}$ is an orthonormal matrix (i.e. $\mathbf{U}^\top\mathbf{U} = \mathbf{I}$) and $\mathbf{D} \in \mathbb{R}^{T\times T}$ is a diagonal matrix. For $\rho > 0$, the optimal solution of the* Tikhonov-regularized problem *in eq.(3) is given by:*

$$\widehat{\mathbf{\Omega}} = \mathbf{U}\widetilde{\mathbf{D}}\mathbf{U}^\top + \frac{1}{\rho}\mathbf{I} \tag{9}$$

*where $\widetilde{\mathbf{D}} \in \mathbb{R}^{T\times T}$ is a diagonal matrix with entries $(\forall t)$ $\widetilde{d}_{tt} = \frac{-d_{tt}}{\rho(d_{tt}+\rho)}$.*

*Proof Sketch.* We consider an *over-complete* singular value decomposition by adding columns to $\mathbf{U}$, thus obtaining $\bar{\mathbf{U}} \in \mathbb{R}^{N\times N}$ such that $\bar{\mathbf{U}}\bar{\mathbf{U}}^\top = \mathbf{I}$, which allows obtaining a singular value decomposition for $\widehat{\mathbf{\Omega}}$. The final result follows from eq.(4). $\square$

Note that computing $\widehat{\mathbf{\Sigma}} \in \mathbb{R}^{N\times N}$ is $\mathcal{O}(N^2T)$-time. In fact, in order to solve the Riccati-regularized problem as well as the Tikhonov-regularized problem, we do not need to compute $\widehat{\mathbf{\Sigma}}$ but its singular value decomposition. The following remark shows that we can obtain the singular value decomposition of $\widehat{\mathbf{\Sigma}}$ by using the singular value decomposition of the training data, which is $\mathcal{O}(NT^2)$-time.

**Remark 4.** *Let $\mathbf{X} \in \mathbb{R}^{N\times T}$ be a dataset composed of $T$ samples, i.e. $\mathbf{X} = (\mathbf{x}^{(1)}, \dots, \mathbf{x}^{(T)})$. Without loss of generality, let assume $\mathbf{X}$ has zero mean, i.e. $\widehat{\boldsymbol{\mu}} = \frac{1}{T}\sum_t \mathbf{x}^{(t)} = \mathbf{0}$. The sample covariance matrix is given by $\widehat{\mathbf{\Sigma}} = \frac{1}{T}\mathbf{X}\mathbf{X}^\top \succeq \mathbf{0}$. We can obtain the singular value decomposition of $\widehat{\mathbf{\Sigma}}$ by using the singular value decomposition of $\mathbf{X}$. That is, $\mathbf{X} = \mathbf{U}\widetilde{\mathbf{D}}\mathbf{V}^\top$ where $\mathbf{U}, \mathbf{V} \in \mathbb{R}^{N\times T}$ are orthonormal matrices (i.e. $\mathbf{U}^\top\mathbf{U} = \mathbf{V}^\top\mathbf{V} = \mathbf{I}$) and $\widetilde{\mathbf{D}} \in \mathbb{R}^{T\times T}$ is a diagonal matrix. We can recover $\mathbf{D}$ from $\widetilde{\mathbf{D}}$ by using $\mathbf{D} = \frac{1}{T}\widetilde{\mathbf{D}}^2$. Finally, we have $\widehat{\mathbf{\Sigma}} = \mathbf{U}\mathbf{D}\mathbf{U}^\top$.*

The following remark shows that instead of obtaining a single solution for a specific value of the regularization parameter $\rho$, we can obtain the whole solution path.

**Remark 5.** *Note that the optimal solutions of the Riccati-regularized problem as well as the Tikhonov-regularized problem have the form $\mathbf{U}\widetilde{\mathbf{D}}\mathbf{U}^\top + c\mathbf{I}$. In these solutions, the diagonal matrix $\widetilde{\mathbf{D}}$ is a function of $\mathbf{D}$ and the regularization parameter $\rho$, while $c$ is a function of $\rho$. Therefore, we need to apply the singular value decomposition only once in order to produce a solution for any value of $\rho$. Furthermore, producing each of those solutions is $\mathcal{O}(T)$-time given that $\widetilde{\mathbf{D}} \in \mathbb{R}^{T\times T}$ is a diagonal matrix.*

### 4.3 Bounds for the Eigenvalues

Here we show that the optimal solution of the Riccati-regularized problem is well defined (i.e. positive definite with finite eigenvalues).

**Corollary 6.** *For $\rho > 0$, the optimal solution of the* Riccati-regularized problem *in eq.(6) is bounded as follows:*

$$\alpha\mathbf{I} \preceq \widehat{\mathbf{\Omega}} \preceq \beta\mathbf{I} \tag{10}$$

*where $\alpha = \sqrt{\frac{1}{\rho} + \frac{1}{4\rho^2}\|\widehat{\mathbf{\Sigma}}\|_2^2} - \frac{1}{2\rho}\|\widehat{\mathbf{\Sigma}}\|_2$ and $\beta = \frac{1}{\sqrt{\rho}}$. That is, the eigenvalues of the optimal solution $\widehat{\mathbf{\Omega}}$ are in the range $[\alpha; \beta]$ and therefore, $\widehat{\mathbf{\Omega}}$ is a well defined precision matrix.*

*Proof Sketch.* Theorem 2 gives the eigendecomposition of the solution $\widehat{\mathbf{\Omega}}$. That is, the diagonal $\widetilde{\mathbf{D}} + \frac{1}{\sqrt{\rho}}\mathbf{I}$ contains the eigenvalues of $\widehat{\mathbf{\Omega}}$. Then we bound the values in the diagonal. $\square$

**Remark 7.** *We can obtain bounds for the* Tikhonov-regularized problem *in eq.(3) by arguments similar to the ones in Corollary 6. That is, the eigenvalues of the optimal solution $\widehat{\mathbf{\Omega}}$ are in the range $[\alpha; \beta]$ where $\alpha = \frac{1}{\|\widehat{\mathbf{\Sigma}}\|_2+\rho}$ and $\beta = \frac{1}{\rho}$.*

### 4.4 Logarithmic Sample Complexity

The $\ell_1$ regularization for Gaussian graphical models has a sample complexity that scales logarithmically with respect to the number of variables $N$ (Ravikumar et al., 2011). On the other hand, it is known that in general, $\ell_2^2$ regularization has a worst-case sample complexity that scales linearly with respect to the number of variables, for problems such as classification (Ng, 2004) and compressed sensing (Cohen et al., 2009).

In this section, we prove the logarithmic sample complexity of the Riccati problem under the *spiked covariance model* originally proposed by (Johnstone, 2001). Without loss of generality, we assume zero mean as in (Ravikumar et al., 2011). First we present the generative model.

**Definition 8.** *The* spiked covariance model *for $N$ variables and $K \ll N$ components is a generative model $\mathcal{P}$ parameterized by $(\mathbf{U}, \mathbf{D}, \beta)$, where $\mathbf{U} \in \mathbb{R}^{N\times K}$ is an orthonormal matrix (i.e. $\mathbf{U}^\top\mathbf{U} = \mathbf{I}$), $\mathbf{D} \in \mathbb{R}^{K\times K}$ is a positive diagonal matrix, and $\beta > 0$. A sample $\mathbf{x} \in \mathbb{R}^N$ is generated by first sampling the independent random vectors $\mathbf{y} \in \mathbb{R}^K$ and $\boldsymbol{\xi} \in \mathbb{R}^N$, both* *with uncorrelated sub-Gaussian entries with zero mean and unit variance. Then we generate the sample* $\mathbf{x}$ *as follows:*

$$\mathbf{x} = \mathbf{U}\mathbf{D}^{1/2}\mathbf{y} + \sqrt{\frac{\beta}{N}}\boldsymbol{\xi} \tag{11}$$

*Furthermore, it is easy to verify that:*

$$\boldsymbol{\Sigma} = \mathbb{E}_{\mathcal{P}}[\mathbf{x}\mathbf{x}^\top] = \mathbf{U}\mathbf{D}\mathbf{U}^\top + \frac{\beta}{N}\mathbf{I} \tag{12}$$

(Please, see Appendix B for a detailed proof.)

Note that the above model is not a Gaussian distribution in general. It is Gaussian if and only if the random variables $\mathbf{y}$ and $\boldsymbol{\xi}$ are Gaussians.

Next we present a concentration inequality for the approximation of the ground truth covariance.

**Lemma 9.** *Let* $\mathcal{P}$ *be a ground truth* spiked covariance model *for* $N$ *variables and* $K \ll N$ *components, parameterized by* $(\mathbf{U}^*, \mathbf{D}^*, \beta^*)$ *with covariance* $\boldsymbol{\Sigma}^*$ *as in Definition 8. Given a sample covariance matrix* $\widehat{\boldsymbol{\Sigma}}$ *computed from* $T$ *samples* $\mathbf{x}^{(1)}, \ldots, \mathbf{x}^{(T)} \in \mathbb{R}^N$ *drawn independently from* $\mathcal{P}$*. With probability at least* $1 - \delta$*, the sample covariance fulfills the following concentration inequality:*

$$\|\widehat{\boldsymbol{\Sigma}} - \boldsymbol{\Sigma}^*\|_{\mathfrak{F}} \leq 40(K\sqrt{\|\mathbf{D}^*\|_{\mathfrak{F}}} + \sqrt{\beta^*})^2 \times \sqrt{\frac{4\log(N+K) + 2\log\frac{4}{\delta}}{T}} \tag{13}$$

*Proof Sketch.* By the definition of the sample covariance, matrix norm inequalities and a concentration inequality for the $\ell_\infty$-norm of a sample covariance matrix from sub-Gaussian random variables given by Lemma 1 in (Ravikumar et al., 2011). □

Armed with the previous result, we present our logarithmic sample complexity for the Riccati problem under the *spiked covariance model.*

**Theorem 10.** *Let* $\mathcal{P}$ *be a ground truth* spiked covariance model *for* $N$ *variables and* $K \ll N$ *components, parameterized by* $(\mathbf{U}^*, \mathbf{D}^*, \beta^*)$ *with covariance* $\boldsymbol{\Sigma}^*$ *as in Definition 8. Let* $\mathcal{Q}^*$ *be the distribution generated by a Gaussian graphical model with zero mean and precision matrix* $\boldsymbol{\Omega}^* = \boldsymbol{\Sigma}^{*-1}$*. That is,* $\mathcal{Q}^*$ *is the projection of* $\mathcal{P}$ *to the family of Gaussian distributions. Given a sample covariance matrix* $\widehat{\boldsymbol{\Sigma}}$ *computed from* $T$ *samples* $\mathbf{x}^{(1)}, \ldots, \mathbf{x}^{(T)} \in \mathbb{R}^N$ *drawn independently from* $\mathcal{P}$*. Let* $\widehat{\mathcal{Q}}$ *be the distribution generated by a Gaussian graphical model with zero mean and precision matrix* $\widehat{\boldsymbol{\Omega}}$*, the solution of the* Riccati-regularized problem *in eq.*(6)*. With probability at least* $1 - \delta$*, we have:*

$$\begin{aligned}&\mathcal{KL}(\mathcal{P}||\widehat{\mathcal{Q}}) - \mathcal{KL}(\mathcal{P}||\mathcal{Q}^*) = \mathbb{E}_{\mathcal{P}}[\mathcal{L}(\boldsymbol{\Omega}^*, \mathbf{x}) - \mathcal{L}(\widehat{\boldsymbol{\Omega}}, \mathbf{x})] \\ &\leq 40(K\sqrt{\|\mathbf{D}^*\|_{\mathfrak{F}}} + \sqrt{\beta^*})^2 \sqrt{\frac{4\log(N+K) + 2\log\frac{4}{\delta}}{T}} \times \\ &\quad \left(\tfrac{1}{4} + \|\boldsymbol{\Omega}^*\|_{\mathfrak{F}} + \|\boldsymbol{\Omega}^*\|_{\mathfrak{F}}^2\right)\end{aligned} \tag{14}$$

*where* $\mathcal{L}(\boldsymbol{\Omega}, \mathbf{x})$ *is the Gaussian log-likelihood as defined in eq.*(1)*.*

*Proof.* Let $p(\mathbf{x})$, $\widehat{q}(\mathbf{x})$ and $q^*(\mathbf{x})$ be the probability density functions of $\mathcal{P}$, $\widehat{\mathcal{Q}}$ and $\mathcal{Q}^*$ respectively. Note that $\mathcal{KL}(\mathcal{P}||\widehat{\mathcal{Q}}) = \mathbb{E}_{\mathcal{P}}[\log p(\mathbf{x}) - \log \widehat{q}(\mathbf{x})] = -\mathcal{H}(\mathcal{P}) - \mathbb{E}_{\mathcal{P}}[\mathcal{L}(\widehat{\boldsymbol{\Omega}}, \mathbf{x})]$. Similarly, $\mathcal{KL}(\mathcal{P}||\mathcal{Q}^*) = -\mathcal{H}(\mathcal{P}) - \mathbb{E}_{\mathcal{P}}[\mathcal{L}(\boldsymbol{\Omega}^*, \mathbf{x})]$. Therefore, $\mathcal{KL}(\mathcal{P}||\widehat{\mathcal{Q}}) - \mathcal{KL}(\mathcal{P}||\mathcal{Q}^*) = \mathbb{E}_{\mathcal{P}}[\mathcal{L}(\boldsymbol{\Omega}^*, \mathbf{x}) - \mathcal{L}(\widehat{\boldsymbol{\Omega}}, \mathbf{x})]$.

By using the definition of the Gaussian log-likelihood in eq.(1), the expected log-likelihood is given by:

$$\mathcal{L}^*(\boldsymbol{\Omega}) \equiv \mathbb{E}_{\mathcal{P}}[\mathcal{L}(\boldsymbol{\Omega}, \mathbf{x})] = \log\det\boldsymbol{\Omega} - \langle \boldsymbol{\Sigma}^*, \boldsymbol{\Omega}\rangle$$

Similarly, define the shorthand notation for the finite-sample log-likelihood in eq.(2) as follows:

$$\widehat{\mathcal{L}}(\boldsymbol{\Omega}) \equiv \mathcal{L}(\boldsymbol{\Omega}, \widehat{\boldsymbol{\Sigma}}) = \log\det\boldsymbol{\Omega} - \langle \widehat{\boldsymbol{\Sigma}}, \boldsymbol{\Omega}\rangle$$

From the above definitions we have:

$$\mathcal{L}^*(\boldsymbol{\Omega}) - \widehat{\mathcal{L}}(\boldsymbol{\Omega}) = \langle \widehat{\boldsymbol{\Sigma}} - \boldsymbol{\Sigma}^*, \boldsymbol{\Omega}\rangle$$

By the Cauchy-Schwarz inequality and Lemma 9:

$$\begin{aligned}|\mathcal{L}^*(\boldsymbol{\Omega}) - \widehat{\mathcal{L}}(\boldsymbol{\Omega})| &\leq \|\widehat{\boldsymbol{\Sigma}} - \boldsymbol{\Sigma}^*\|_{\mathfrak{F}}\|\boldsymbol{\Omega}\|_{\mathfrak{F}} \\ &\leq \gamma\|\boldsymbol{\Omega}\|_{\mathfrak{F}}\end{aligned}$$

for $\gamma = 40(K\sqrt{\|\mathbf{D}^*\|_{\mathfrak{F}}} + \sqrt{\beta^*})^2\sqrt{\frac{4\log(N+K)+2\log\frac{4}{\delta}}{T}}$. Thus, we obtained a "uniform convergence" statement for the loss.

Note that $\widehat{\boldsymbol{\Omega}}$ (the minimizer of the regularized finite-sample loss) and $\boldsymbol{\Omega}^*$ (the minimizer of the expected loss) fulfill by definition:

$$\begin{aligned}\widehat{\boldsymbol{\Omega}} &= \arg\max_{\boldsymbol{\Omega}} \widehat{\mathcal{L}}(\boldsymbol{\Omega}) - \frac{\rho}{2}\|\boldsymbol{\Omega}\|_{\mathfrak{F}}^2 \\ \boldsymbol{\Omega}^* &= \arg\max_{\boldsymbol{\Omega}} \mathcal{L}^*(\boldsymbol{\Omega})\end{aligned}$$

Therefore, by optimality of $\widehat{\boldsymbol{\Omega}}$ we have:

$$\begin{aligned}&\widehat{\mathcal{L}}(\boldsymbol{\Omega}^*) - \frac{\rho}{2}\|\boldsymbol{\Omega}^*\|_{\mathfrak{F}}^2 \leq \widehat{\mathcal{L}}(\widehat{\boldsymbol{\Omega}}) - \frac{\rho}{2}\|\widehat{\boldsymbol{\Omega}}\|_{\mathfrak{F}}^2 \\ \Rightarrow\ &\widehat{\mathcal{L}}(\boldsymbol{\Omega}^*) - \widehat{\mathcal{L}}(\widehat{\boldsymbol{\Omega}}) \leq \frac{\rho}{2}\|\boldsymbol{\Omega}^*\|_{\mathfrak{F}}^2 - \frac{\rho}{2}\|\widehat{\boldsymbol{\Omega}}\|_{\mathfrak{F}}^2\end{aligned}$$

By using the "uniform convergence" statement, the above results and setting $\rho = 2\gamma$:

$$\begin{aligned}\mathcal{L}^*(\boldsymbol{\Omega}^*) - \mathcal{L}^*(\widehat{\boldsymbol{\Omega}}) &\leq \widehat{\mathcal{L}}(\boldsymbol{\Omega}^*) - \widehat{\mathcal{L}}(\widehat{\boldsymbol{\Omega}}) + \gamma\|\boldsymbol{\Omega}^*\|_{\mathfrak{F}} + \gamma\|\widehat{\boldsymbol{\Omega}}\|_{\mathfrak{F}} \\ &\leq \frac{\rho}{2}\|\boldsymbol{\Omega}^*\|_{\mathfrak{F}}^2 - \frac{\rho}{2}\|\widehat{\boldsymbol{\Omega}}\|_{\mathfrak{F}}^2 + \gamma\|\boldsymbol{\Omega}^*\|_{\mathfrak{F}} + \gamma\|\widehat{\boldsymbol{\Omega}}\|_{\mathfrak{F}} \\ &= \gamma(\|\widehat{\boldsymbol{\Omega}}\|_{\mathfrak{F}} - \|\widehat{\boldsymbol{\Omega}}\|_{\mathfrak{F}}^2 + \|\boldsymbol{\Omega}^*\|_{\mathfrak{F}} + \|\boldsymbol{\Omega}^*\|_{\mathfrak{F}}^2)\end{aligned}$$

By noting that $\|\widehat{\mathbf{\Omega}}\|_{\mathfrak{F}} - \|\widehat{\mathbf{\Omega}}\|_{\mathfrak{F}}^2 \leq \frac{1}{4}$, we prove our claim. □

### 4.5 Spectral Method for the Sparse Problem

In this section, we propose sparsification of the dense solution and show an upper bound of the degradation of the Kullback-Leibler divergence and expected log-likelihood.

First, we study the relationship between the sparseness of the low-rank parameterization and the generated precision matrix. We analyze the expected value of densities of random matrices. We assume that each entry in the matrix is statistically independent, which is not a novel assumption. From a Bayesian viewpoint, the $\ell_1$-regularized problem in eq.(5) assumes that each entry in the precision matrix $\mathbf{\Omega}$ is independent. We also made a similar assumption for our Riccati-regularized problem in eq.(6).

**Lemma 11.** *Let $\mathbf{A} \in \mathbb{R}^{N\times T}$ be a random matrix with statistically independent entries and expected density $p$, i.e. $(\forall n,t)\ \mathbb{P}[a_{nt} \neq 0] = p$. Furthermore, assume that conditional distribution of $a_{nt} \mid a_{nt} \neq 0$ has a domain with non-zero Lebesgue measure. Let $\mathbf{D} \in \mathbb{R}^{T\times T}$ be a diagonal matrix such that $(\forall t)\ d_{tt} \neq 0$. Let $c$ be an arbitrary real value. Let $\mathbf{B} = \mathbf{ADA}^\top + c\mathbf{I}$. The expected non-diagonal density of $\mathbf{B} \in \mathbb{R}^{N\times N}$ is given by:*

$$(\forall n_1 \neq n_2)\ \mathbb{P}[b_{n_1n_2} \neq 0] = 1 - (1-p^2)^T \qquad (15)$$

*Proof Sketch.* Note that for $n_1 \neq n_2$, we have $b_{n_1n_2} = \sum_t d_{tt}a_{n_1t}a_{n_2t}$. We argue that the probability that $b_{n_1n_2} \neq 0$ is equal to the probability that $a_{n_1t} \neq 0$ and $a_{n_2t} \neq 0$ for at least one $t$. The final result follows from independence of the entries of $\mathbf{A}$. □

It is easy to construct (non-random) specific instances in which the density of $\mathbf{A}$ and $\mathbf{B}$ are unrelated. Consider two examples in which $\mathbf{D} = \mathbf{I}$, $c = 0$ and therefore $\mathbf{B} = \mathbf{AA}^\top$. As a first example, let $\mathbf{A}$ have zero entries except for $(\forall n)\ a_{n1} = 1$. That is, $\mathbf{A}$ is very sparse but $\mathbf{B} = \mathbf{AA}^\top = \mathbf{11}^\top$ is dense. As a second example, let $\mathbf{B}$ have zero entries except for $(\forall n)\ b_{n1} = b_{1n} = 1, b_{nn} = N$. That is, $\mathbf{B} = \mathbf{AA}^\top$ is very sparse but its Chokesly decomposition $\mathbf{A}$ is dense.

Next, we show that sparsification of a dense precision matrix produces a precision matrix that is close to the original one. Furthermore, we show that such matrix is positive definite.

**Theorem 12.** *Let $\mathbf{\Omega} = \mathbf{UDU}^\top + \beta\mathbf{I}$ be a precision matrix, where $\mathbf{U} \in \mathbb{R}^{N\times T}$ is an orthonormal matrix (i.e. $\mathbf{U}^\top\mathbf{U} = \mathbf{I}$), $\beta > \alpha > 0$ and $\mathbf{D} \in \mathbb{R}^{T\times T}$ is a negative diagonal matrix such that $(\forall t)\ -(\beta-\alpha) < d_{tt} \leq 0$ (or equivalently $\alpha\mathbf{I} \preceq \mathbf{\Omega} \preceq \beta\mathbf{I}$). Let $\widetilde{\mathbf{U}} \in \mathbb{R}^{N\times T}$ be a sparse matrix constructed by soft-thresholding:*

$$(\forall n,t)\ \widetilde{u}_{nt} = \operatorname{sign}(u_{nt}) \max\left(0, |u_{nt}| - \tfrac{\lambda}{\sqrt{NT}}\right) \qquad (16)$$

*or by hard-thresholding:*

$$(\forall n,t)\ \widetilde{u}_{nt} = u_{nt}1\left[|u_{nt}| \geq \tfrac{\lambda}{\sqrt{NT}}\right] \qquad (17)$$

*for some $\lambda > 0$. The* sparse *precision matrix $\widetilde{\mathbf{\Omega}} = \widetilde{\mathbf{U}}\mathbf{D}\widetilde{\mathbf{U}}^\top + \beta\mathbf{I}$ satisfies:*

$$\|\widetilde{\mathbf{\Omega}} - \mathbf{\Omega}\|_2 \leq (2\lambda + \lambda^2)(\beta - \alpha) \qquad (18)$$

*Furthermore, $\widetilde{\mathbf{\Omega}}$ preserves positive definiteness of the* dense *precision matrix $\mathbf{\Omega}$. More formally:*

$$\alpha\mathbf{I} \preceq \widetilde{\mathbf{\Omega}} \preceq \beta\mathbf{I} \qquad (19)$$

*That is, the eigenvalues of $\widetilde{\mathbf{\Omega}}$ are in the range $[\alpha;\beta]$ and therefore, $\widetilde{\mathbf{\Omega}}$ is a well defined* sparse *precision matrix.*

*Proof Sketch.* The claims in eq.(18) and eq.(19) follow from matrix norm inequalities. Additionally, eq.(19) follows from showing that $\widetilde{\mathbf{U}}\mathbf{D}\widetilde{\mathbf{U}}^\top$ is negative semidefinite and that $\|\widetilde{\mathbf{U}}\|_2 \leq \|\mathbf{U}\|_2 = 1$. □

Finally, we derive an upper bound of the degradation of the Kullback-Leibler divergence and expected log-likelihood. Thus, sparsification of a dense solution has provable performance guarantees.

**Theorem 13.** *Let $\mathcal{P}$ be the ground truth distribution of the data sample $\mathbf{x}$. Let $\mathcal{Q}$ be the distribution generated by a Gaussian graphical model with mean $\boldsymbol{\mu}$ and precision matrix $\mathbf{\Omega}$. Let $\widetilde{\mathcal{Q}}$ be the distribution generated by a Gaussian graphical model with mean $\boldsymbol{\mu}$ and precision matrix $\widetilde{\mathbf{\Omega}}$. Assume $\alpha\mathbf{I} \preceq \mathbf{\Omega}$ and $\alpha\mathbf{I} \preceq \widetilde{\mathbf{\Omega}}$ for $\alpha > 0$. The Kullback-Leibler divergence from $\mathcal{P}$ to $\widetilde{\mathcal{Q}}$ is close to the Kullback-Leibler divergence from $\mathcal{P}$ to $\mathcal{Q}$. More formally:*

$$\begin{aligned}\mathcal{KL}(\mathcal{P}\|\widetilde{\mathcal{Q}}) - \mathcal{KL}(\mathcal{P}\|\mathcal{Q}) &= \mathbb{E}_{\mathcal{P}}[\mathcal{L}(\mathbf{\Omega},\mathbf{x}) - \mathcal{L}(\widetilde{\mathbf{\Omega}},\mathbf{x})] \\ &\leq \left(\frac{1}{\alpha} + \|\mathbb{E}_{\mathcal{P}}[(\mathbf{x}-\boldsymbol{\mu})(\mathbf{x}-\boldsymbol{\mu})^\top]\|_2\right)\|\widetilde{\mathbf{\Omega}} - \mathbf{\Omega}\|_2 \qquad (20)\end{aligned}$$

*where $\mathcal{L}(\mathbf{\Omega},\mathbf{x})$ is the Gaussian log-likelihood as defined in eq.(1). Moreover, the above bound is finite provided that $\|\mathbb{E}_{\mathcal{P}}[\mathbf{x}]\|_2$ and $\|\mathbb{E}_{\mathcal{P}}[\mathbf{xx}^\top]\|_2$ are finite.*

*Proof Sketch.* By definition of the Kullback-Leibler divergence, the Gaussian log-likelihood in eq.(1) and by showing that the resulting expression is Lipschitz continuous with respect to the spectral norm. □

### 4.6 Using the Learnt Models in Linear-Time

Next we provide techniques for using our learnt models, such as removing unimportant variables, computing likelihoods and conditional distributions.

**Removal of Unimportant Variables.** Recall that the final goal for a data analyst is not only to be able to learn models from data but also to browse such learnt models in order to discover "important" interactions. We define the *unimportant variable set* as the set of variables in which the absolute value of every *partial correlation* is lower than a specified threshold.

**Definition 14.** *An* unimportant variable set *of a Gaussian graphical model defined by the precision matrix $\mathbf{\Omega} = \{\omega_{n_1 n_2}\}$ for threshold $\varepsilon > 0$ is a set $\mathcal{S} \subseteq \{1, \ldots, N\}$ such that:*

$$(\forall n_1, n_2 \in \mathcal{S}) \ \frac{|\omega_{n_1 n_2}|}{\sqrt{\omega_{n_1 n_1}\omega_{n_2 n_2}}} \leq \varepsilon \tag{21}$$

Note that our definition does not require that the size of $\mathcal{S}$ is maximal. Here we provide a technique that discovers the *unimportant variable set* in linear-time.

**Lemma 15.** *Let $\mathbf{\Omega} = \mathbf{A}\mathbf{D}\mathbf{A}^\top + c\mathbf{I}$ be a precision matrix, where $\mathbf{A} \in \mathbb{R}^{N\times T}$ is an arbitrary matrix, $\mathbf{D} \in \mathbb{R}^{T\times T}$ is an arbitrary diagonal matrix, and $c$ is an arbitrary real value, such that $\mathbf{\Omega}$ is a well defined precision matrix (i.e. $\mathbf{\Omega} \succ \mathbf{0}$). An* unimportant variable set *of the Gaussian graphical model defined by $\mathbf{\Omega} = \{\omega_{n_1 n_2}\}$ for threshold $\varepsilon > 0$ can be detected by applying the rule:*

$$(\forall n_1, n_2 \in \mathcal{S}) \ \frac{|\omega_{n_1 n_2}|}{\sqrt{\omega_{n_1 n_1}\omega_{n_2 n_2}}} \leq \min\left(q(n_1), q(n_2)\right) \leq \varepsilon \tag{22}$$

*where $q(n_1) = \frac{\sum_t |d_{tt} a_{n_1 t}| \max_{n_2} |a_{n_2 t}|}{\sqrt{r(n_1) \min_{n_2} r(n_2)}}$ and $r(n) = \sum_t d_{tt} a_{nt}^2 + c$. Furthermore, this operation has time-complexity $\mathcal{O}(NT)$ when applied to all variables in $\{1, \ldots, N\}$.*

*Proof Sketch.* Straightforward, from algebra. □

**Computation of the Likelihood.** In tasks such as classification, we would assign a given sample to the class whose model produced the highest likelihood among all classes. Here we provide a method for computing the likelihood of a given sample in the learnt model.

**Lemma 16.** *Let $\mathbf{\Omega} = \mathbf{A}\mathbf{D}\mathbf{A}^\top + c\mathbf{I}$ be a precision matrix, where $\mathbf{A} \in \mathbb{R}^{N\times T}$ is an arbitrary matrix, $\mathbf{D} \in \mathbb{R}^{T\times T}$ is an arbitrary diagonal matrix, and $c$ is an arbitrary real value, such that $\mathbf{\Omega}$ is a well defined precision matrix (i.e. $\mathbf{\Omega} \succ \mathbf{0}$). The log-likelihood of a sample $\mathbf{x}$ in a Gaussian graphical model with mean $\boldsymbol{\mu}$ and precision matrix $\mathbf{\Omega}$ is given by:*

$$\begin{aligned}\mathcal{L}(\mathbf{\Omega}, \mathbf{x}) = &\log\det\left(\mathbf{I} + \frac{1}{c}\mathbf{A}^\top\mathbf{A}\mathbf{D}\right) + N\log c \\ &- ((\mathbf{x}-\boldsymbol{\mu})^\top\mathbf{A})\mathbf{D}(\mathbf{A}^\top(\mathbf{x}-\boldsymbol{\mu})) \\ &- c(\mathbf{x}-\boldsymbol{\mu})^\top(\mathbf{x}-\boldsymbol{\mu})\end{aligned} \tag{23}$$

*where $\mathcal{L}(\mathbf{\Omega}, \mathbf{x})$ is the Gaussian log-likelihood as defined in eq.(1). Furthermore, this operation has time-complexity $\mathcal{O}(NT^2)$.*

*Proof Sketch.* By the matrix determinant lemma. □

**Conditional Distributions.** In some contexts, it is important to perform inference for some unobserved variables when given some observed variables. Here we provide a method for computing the conditional distribution that takes advantage of our low-rank parameterization. First, we show how to transform from a non-orthonormal parameterization to an orthonormal one.

**Remark 17.** *Let $\mathbf{A} \in \mathbb{R}^{N\times T}$ be an arbitrary matrix. Let $\mathbf{D} \in \mathbb{R}^{T\times T}$ be a negative diagonal matrix (i.e. $(\forall t)\ d_{tt} < 0$). Let $\mathbf{B} = \mathbf{A}\mathbf{D}\mathbf{A}^\top$. Since $\mathbf{B}$ has rank at most $T$, we can compute its singular value decomposition $\mathbf{B} = \mathbf{U}\widetilde{\mathbf{D}}\mathbf{U}^\top$ where $\mathbf{U} \in \mathbb{R}^{N\times T}$ is an orthonormal matrix (i.e. $\mathbf{U}^\top\mathbf{U} = \mathbf{I}$) and $\widetilde{\mathbf{D}} \in \mathbb{R}^{T\times T}$ is a negative diagonal matrix (i.e. $(\forall t)\ \widetilde{d}_{tt} < 0$). In order to do this, we compute the singular value decomposition of $\mathbf{A}(-\mathbf{D})^{1/2} = \mathbf{U}\widetilde{\mathbf{D}}\mathbf{V}^\top$. We can recover $\mathbf{D}$ from $\widetilde{\mathbf{D}}$ by using $\mathbf{D} = -\widetilde{\mathbf{D}}^2$.*

Given the previous observation, our computation of conditional distributions will focus only on orthonormal parameterizations.

**Lemma 18.** *Let $\mathbf{\Omega} = \mathbf{U}\mathbf{D}\mathbf{U}^\top + c\mathbf{I}$ be a precision matrix, where $\mathbf{U} \in \mathbb{R}^{N\times T}$ is an orthonormal matrix (i.e. $\mathbf{U}^\top\mathbf{U} = \mathbf{I}$), $\mathbf{D} \in \mathbb{R}^{T\times T}$ is an arbitrary diagonal matrix, and $c$ is an arbitrary real value, such that $\mathbf{\Omega}$ is a well defined precision matrix (i.e. $\mathbf{\Omega} \succ \mathbf{0}$). Assume that $\mathbf{x}$ follows a Gaussian graphical model with mean $\boldsymbol{\mu}$ and precision matrix $\mathbf{\Omega}$, and assume that we partition the variables into two sets as follows:*

$$\mathbf{x} = \begin{bmatrix}\mathbf{x}_1 \\ \mathbf{x}_2\end{bmatrix}, \boldsymbol{\mu} = \begin{bmatrix}\boldsymbol{\mu}_1 \\ \boldsymbol{\mu}_2\end{bmatrix}, \mathbf{U} = \begin{bmatrix}\mathbf{U}_1 \\ \mathbf{U}_2\end{bmatrix} \tag{24}$$

*The conditional distribution of $\mathbf{x}_1 \mid \mathbf{x}_2$ is a Gaussian graphical model with mean $\boldsymbol{\mu}_{1|2}$ and precision matrix $\mathbf{\Omega}_{11}$, such that:*

$$\begin{aligned}&\text{i. } \boldsymbol{\mu}_{1|2} = \boldsymbol{\mu}_1 - \mathbf{U}_1\widetilde{\mathbf{D}}\mathbf{U}_2^\top(\mathbf{x}_2 - \boldsymbol{\mu}_2) \\ &\text{ii. } \mathbf{\Omega}_{11} = \mathbf{U}_1\mathbf{D}\mathbf{U}_1^\top + c\mathbf{I}\end{aligned} \tag{25}$$

*where $\widetilde{\mathbf{D}} \in \mathbb{R}^{T\times T}$ is a diagonal matrix with entries $(\forall t)\ \widetilde{d}_{tt} = \frac{d_{tt}}{d_{tt}+c}$.*

*Proof Sketch.* By definition of conditional distributions of Gaussians (Lauritzen, 1996) and Theorem 3. □

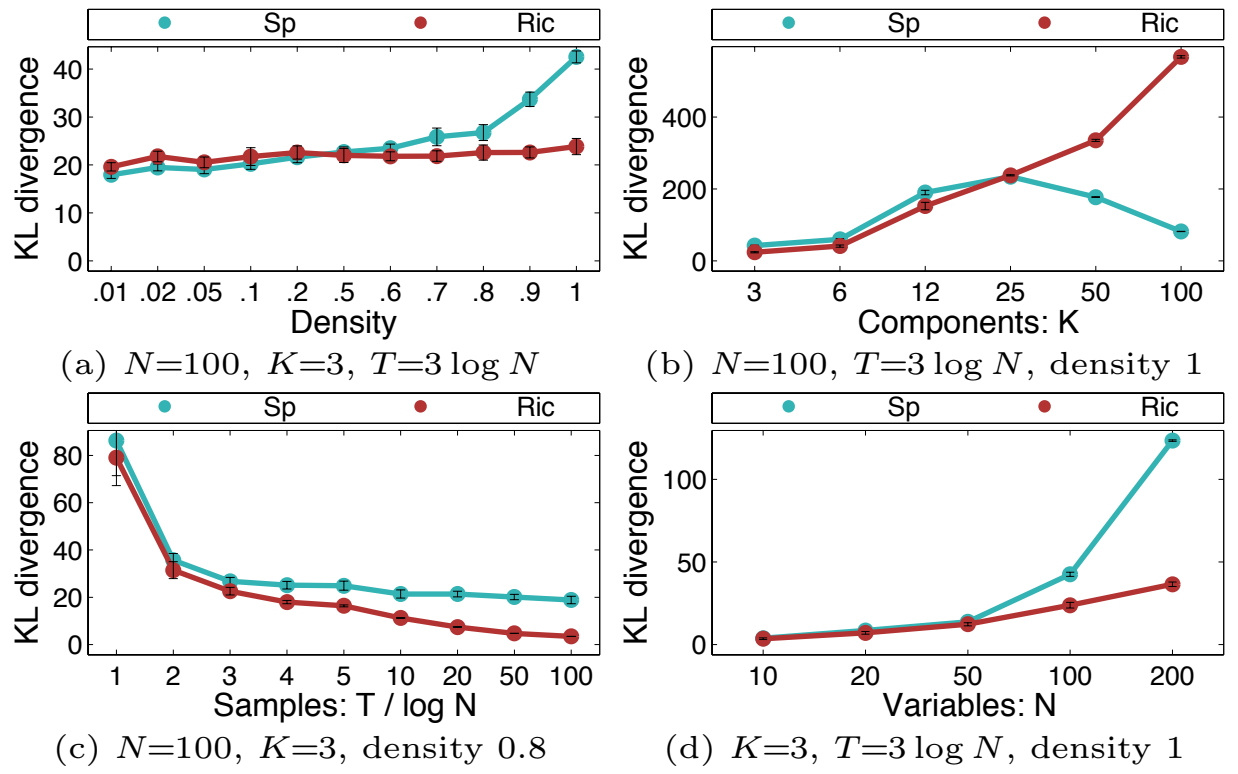

(a) $N$=100, $K$=3, $T$=3 log $N$ (b) $N$=100, $T$=3 log $N$, density 1

(c) $N$=100, $K$=3, density 0.8 (d) $K$=3, $T$=3 log $N$, density 1

Figure 1: Kullback-Leibler divergence between the ground truth *spiked covariance model* and the learnt models for different (a) inverse covariance densities, (b) number of components, (c) samples and (d) variables (error bars shown at 90% confidence interval). The Riccati (Ric) method performs better for ground truth with moderate to high density, while the sparse method (Sp) performs better for low densities. The sparse method performs better for ground truth with large number of components, while the Riccati method performs better for small number of components. The behavior of both methods is similar with respect to the number of samples. The sparse method degrades more than the Riccati method with respect to the number of variables.

Table 2: Gene expression datasets.

| **Dataset** | **Disease** | **Samples** | **Variables** |
|---|---|---|---|
| GSE1898 | Liver cancer | 182 | 21,794 |
| GSE29638 | Colon cancer | 50 | 22,011 |
| GSE30378 | Colon cancer | 95 | 22,011 |
| GSE20194 | Breast cancer | 278 | 22,283 |
| GSE22219 | Breast cancer | 216 | 24,332 |
| GSE13294 | Colon cancer | 155 | 54,675 |
| GSE17951 | Prostate cancer | 154 | 54,675 |
| GSE18105 | Colon cancer | 111 | 54,675 |
| GSE1476 | Colon cancer | 150 | 59,381 |
| GSE14322 | Liver cancer | 76 | 104,702 |
| GSE18638 | Colon cancer | 98 | 235,826 |
| GSE33011 | Ovarian cancer | 80 | 367,657 |
| GSE30217 | Leukemia | 54 | 964,431 |
| GSE33848 | Lung cancer | 30 | 1,852,426 |

Table 3: Runtimes for gene expression datasets. Our Riccati method is considerably faster than sparse method.

| **Dataset** | **Sparse** | **Riccati** |
|---|---|---|
| GSE1898,29638,30378,20194,22219 | 3.8 min | 1.2 sec |
| GSE13294,17951,18105,1476 | 14.9 min | 1.6 sec |
| GSE14322 | 30.4 min | 1.0 sec |
| GSE18638 | 3.1 hr | 2.6 sec |
| GSE33011 | 6.0 hr | 2.5 sec |
| GSE30217 | 1.3 days | 3.8 sec |
| GSE33848 | 5.4 days | 2.8 sec |

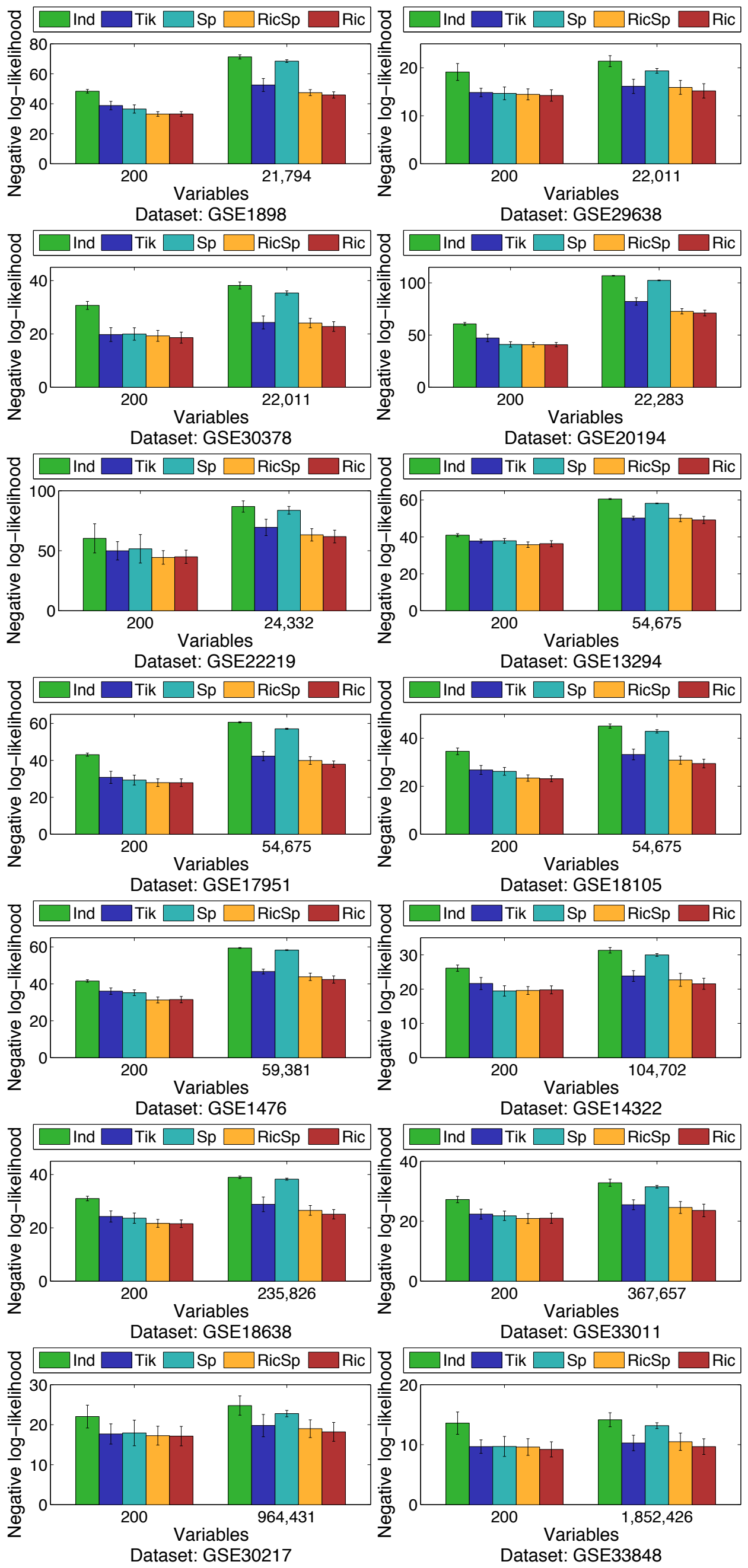

Figure 2: Negative test log-likelihood for gene expression datasets (error bars shown at 90% confidence interval). Our Riccati (Ric) and sparse Riccati (RicSp) methods perform better than the sparse method (Sp), Tikhonov method (Tik) and the fully independent model (Ind). Since the method of (Mazumder & Hastie, 2012) requires high regularization for producing reasonably sized components, the performance of the sparse method (Sp) when using all the variables degrade considerably when compared to 200 variables.

## 5 Experimental Results

We begin with a synthetic example to test the ability of the method to recover the ground truth distribution from data. We used the *spiked covariance model* as in Definition 8 for $N = 100$ variables, $K = 3$ components and $\beta = 1$. Additionally, we control for the density of the related ground truth inverse covariance matrix. We performed 20 repetitions. For each repetition, we generate a different ground truth model at random.

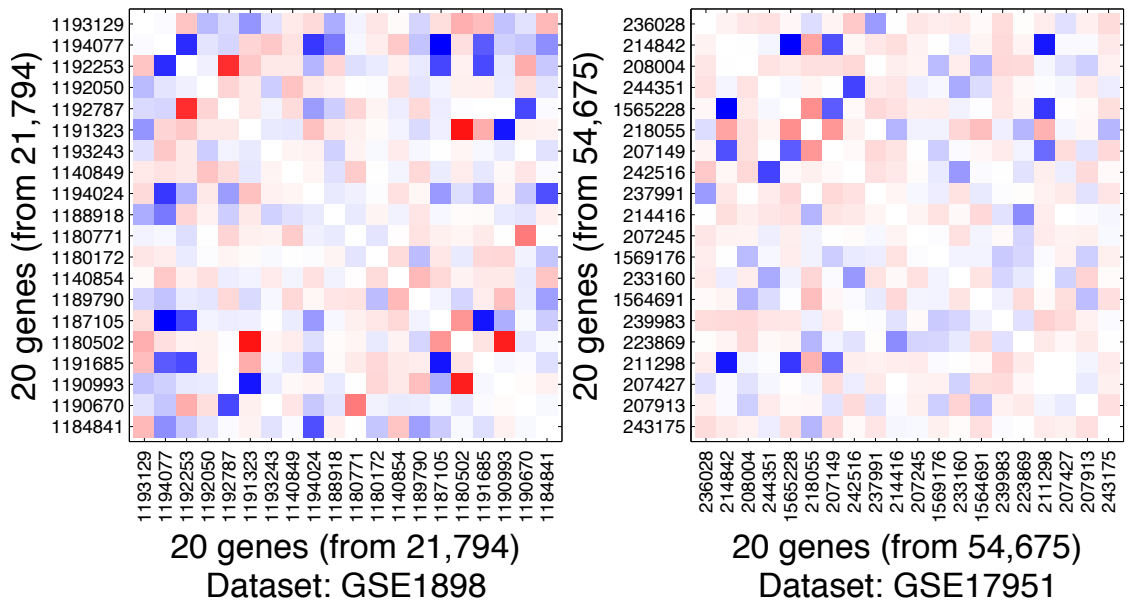

Figure 3: Genes with "important" interactions for two gene expression datasets analyzed with our Riccati method.

We then produce $T = 3 \log N$ samples for training and the same number of samples for validation. In our experiments, we test different scenarios, by varying the density of the ground truth inverse covariance, the number of components, samples and variables.

In Figure 1, we report the Kullback-Leibler divergence between the ground truth and the learnt models. For learning sparse models, we used the method of (Friedman et al., 2007). The Riccati method performs better for ground truth with moderate to high density, while the sparse method performs better for low densities. The sparse method performs better for ground truth with large number of components (as expected when $K \in \mathcal{O}(N)$), while the Riccati method performs better for small number of components (as expected when $K \ll N$). The behavior of both methods is similar with respect to the number of samples. The sparse method degrades more than the Riccati method with respect to the number of variables.

For experimental validation on real-world datasets, we used 14 cancer datasets publicly available at the Gene Expression Omnibus (`http://www.ncbi.nlm.nih.gov/geo/`). Table 2 shows the datasets as well as the number of samples and variables on each of them. We preprocessed the data so that each variable is zero mean and unit variance across the dataset. We performed 50 repetitions. For each repetition, we used one third of the samples for training, one third for validation and the remaining third for testing. We report the negative log-likelihood on the testing set (after subtracting the entropy measured on the testing set in order to make it comparable to the Kullback-Leibler divergence).

Since regular sparseness promoting methods do not scale to our setting of more than 20 thousands variables, we validate our method in two regimes. In the first regime, for each of the 50 repetitions, we select $N = 200$ variables uniformly at random and use the sparse method of (Friedman et al., 2007). In the second regime, we use all the variables in the dataset, and use the method of (Mazumder & Hastie, 2012) so that the biggest connected component has at most $N' = 500$ variables (as prescribed in (Mazumder & Hastie, 2012)). The technique of (Mazumder & Hastie, 2012) computes a graph from edges with an absolute value of the covariance higher than the regularization parameter $\rho$ and then splits the graph into its connected components. Since the whole sample covariance matrix could not fit in memory, we computed it in batches of rows (Mazumder & Hastie, 2012). Unfortunately, in order to make the biggest connected component of a reasonable size $N' = 500$ (as prescribed in (Mazumder & Hastie, 2012)), a high value of $\rho$ is needed. In order to circumvent this problem, we used a high value of $\rho$ for splitting the graph into its connected components, and allowed for low values of $\rho$ for computing the precision matrix for each component. For our sparse Riccati method, we used soft-thresholding of the Riccati solution with $\lambda = \rho$.

In Figure 2, we observe that our Riccati and sparse Riccati methods perform better than the comparison methods. Since the method of (Mazumder & Hastie, 2012) requires high regularization for producing reasonably sized components, the performance of the sparse method when using all the variables degrade considerably when compared to 200 variables.

In Figure 3, we show the interaction between a set of genes that were selected after applying our rule for removing unimportant variables.

Finally, we show average runtimes in Table 3. In order to make a fair comparison, the runtime includes the time needed to produce the optimal precision matrix from a given input dataset. This includes not only the time to solve each optimization problem but also the time to compute the covariance matrix (if needed). Our Riccati method is considerably faster than the sparse method.

## 6 Concluding Remarks

We can generalize our penalizer to one of the form $\|\mathbf{A\Omega}\|_{\mathfrak{F}}^2$ and investigate under which conditions, this problem has a low-rank solution. Another extension includes finding inverse covariance matrices that are both sparse and low-rank. While in this paper we show loss consistency, we could also analyze conditions for which the recovered parameters approximate the ground truth, similar to the work of (Rothman et al., 2008; Ravikumar et al., 2011). Proving consistency of sparse patterns is a challenging line of research, since our low-rank estimators would not have enough degrees of freedom in order to accomodate for all possible sparseness patterns. We conjecture that such results might be possible by borrowing from the literature on principal component analysis (Nadler, 2008).

## References

Banerjee, O., El Ghaoui, L., & d'Aspremont, A. (2008). Model selection through sparse maximum likelihood estimation for multivariate Gaussian or binary data. *JMLR*.

Banerjee, O., El Ghaoui, L., d'Aspremont, A., & Natsoulis, G. (2006). Convex optimization techniques for fitting sparse Gaussian graphical models. *ICML*.

Cohen, A., Dahmen, W., & DeVore, R. (2009). Compressed sensing and best k-term approximation. *J. Amer. Math. Soc.*

Dempster, A. (1972). Covariance selection. *Biometrics*.

Duchi, J., Gould, S., & Koller, D. (2008). Projected subgradient methods for learning sparse Gaussians. *UAI*.

Friedman, J., Hastie, T., & Tibshirani, R. (2007). Sparse inverse covariance estimation with the graphical lasso. *Biostatistics*.

Guillot, D., Rajaratnam, B., Rolfs, B., Maleki, A., & Wong, I. (2012). Iterative thresholding algorithm for sparse inverse covariance estimation. *NIPS*.

Hsieh, C., Dhillon, I., Ravikumar, P., & Banerjee, A. (2012). A divide-and-conquer procedure for sparse inverse covariance estimation. *NIPS*.

Hsieh, C., Sustik, M., Dhillon, I., & Ravikumar, P. (2011). Sparse inverse covariance matrix estimation using quadratic approximation. *NIPS*.

Huang, J., Liu, N., Pourahmadi, M., & Liu, L. (2006). Covariance matrix selection and estimation via penalized normal likelihood. *Biometrika*.

Johnson, C., Jalali, A., & Ravikumar, P. (2012). High-dimensional sparse inverse covariance estimation using greedy methods. *AISTATS*.

Johnstone, I. (2001). On the distribution of the largest principal component. *The Annals of Statistics*.

Lauritzen, S. (1996). *Graphical models*. Oxford Press.

Lim, Y. (2006). The matrix golden mean and its applications to Riccati matrix equations. *SIAM Journal on Matrix Analysis and Applications*.

Lu, Z. (2009). Smooth optimization approach for sparse covariance selection. *SIAM Journal on Optimization*.

Mazumder, R., & Hastie, T. (2012). Exact covariance thresholding into connected components for large-scale graphical lasso. *JMLR*.

Meinshausen, N., & Bühlmann, P. (2006). High dimensional graphs and variable selection with the lasso. *The Annals of Statistics*.

Nadler, B. (2008). Finite sample approximation results for principal component analysis: A matrix perturbation approach. *The Annals of Statistics*.

Ng, A. (2004). Feature selection, $\ell_1$ vs. $\ell_2$ regularization, and rotational invariance. *ICML*.

Olsen, P., Oztoprak, F., Nocedal, J., & Rennie, S. (2012). Newton-like methods for sparse inverse covariance estimation. *NIPS*.

Ravikumar, P., Wainwright, M., Raskutti, G., & Yu, B. (2011). High-dimensional covariance estimation by minimizing $\ell_1$-penalized log-determinant divergence. *Electronic Journal of Statistics*.

Rothman, A., Bickel, P., Levina, E., & Zhu, J. (2008). Sparse permutation invariant covariance estimation. *Electronic Journal of Statistics*.

Scheinberg, K., Ma, S., & Goldfarb, D. (2010). Sparse inverse covariance selection via alternating linearization methods. *NIPS*.

Scheinberg, K., & Rish, I. (2010). Learning sparse Gaussian Markov networks using a greedy coordinate ascent approach. *ECML*.

Schmidt, M., van den Berg, E., Friedlander, M., & Murphy, K. (2009). Optimizing costly functions with simple constraints: A limited-memory projected quasi-Newton algorithm. *AISTATS*.

Witten, D., & Tibshirani, R. (2009). Covariance-regularized regression and classification for high-dimensional problems. *Journal of the Royal Statistical Society*.

Yuan, M., & Lin, Y. (2007). Model selection and estimation in the Gaussian graphical model. *Biometrika*.

Yun, S., Tseng, P., & Toh, K. (2011). A block coordinate gradient descent method for regularized convex separable optimization and covariance selection. *Mathematical Programming*.

Zhou, S., Rütimann, P., Xu, M., & Bühlmann, P. (2011). High-dimensional covariance estimation based on Gaussian graphical models. *JMLR*.